\documentclass{article}

\usepackage[final]{corl_2020} 

\usepackage[utf8]{inputenc} 
\usepackage[T1]{fontenc}    
\usepackage{hyperref}       
\usepackage{url}            
\usepackage{booktabs}       
\usepackage{amsfonts}       
\usepackage{nicefrac}       
\usepackage{microtype}      

\usepackage{graphicx}
\usepackage{amssymb}
\usepackage{amsfonts}
\usepackage{amsmath}
\usepackage{booktabs}
\usepackage{algorithm}
\usepackage{algorithmicx}
\usepackage[noend]{algpseudocode}
\usepackage{wrapfig}
\usepackage{bm}
\usepackage{bbm}
\usepackage{breqn}
\usepackage{natbib}
\usepackage[font=small]{caption}
\usepackage[table]{xcolor}
\usepackage{array,multirow}

\usepackage{todonotes}

\usepackage{epigraph}
\usepackage{mathtools}

\DeclarePairedDelimiterX{\kldivx}[2]{(}{)}{%
  #1\;\delimsize\|\;#2%
}
\newcommand{\kldiv}{D_{\mathrm{KL}}\kldivx}

\DeclareMathOperator*{\argmax}{arg\,max}
\DeclareMathOperator*{\argmin}{arg\,min}

\setcitestyle{authoryear,open={(},close={)}}

\newcommand{\bs}{\mathbf{s}}
\newcommand{\ba}{\mathbf{a}}
\newcommand{\bo}{\mathbf{o}}
\newcommand{\bI}{\mathbf{I}}

\title{Assisted Perception:\\Optimizing Observations to Communicate State}

\author{%
  Siddharth Reddy, Sergey Levine, Anca D. Dragan \\
  Department of Electrical Engineering and Computer Science\\
  University of California, Berkeley\\
  \texttt{\{sgr,svlevine,anca\}@berkeley.edu} \\
}

\begin{document}

\maketitle

\begin{abstract}
We aim to help users estimate the state of the world in tasks like robotic teleoperation and navigation with visual impairments, where users may have systematic biases that lead to suboptimal behavior: they might struggle to process observations from multiple sensors simultaneously, receive delayed observations, or overestimate distances to obstacles.
While we cannot directly change the user's internal beliefs or their internal state estimation process, our insight is that we can still assist them by modifying the user's observations.
Instead of showing the user their true observations, we synthesize new observations that lead to more accurate internal state estimates when processed by the user.
We refer to this method as assistive state estimation (ASE): an automated assistant uses the true observations to infer the state of the world, then generates a modified observation for the user to consume (e.g., through an augmented reality interface), and optimizes the modification to induce the user's new beliefs to match the assistant's current beliefs.
To predict the effect of the modified observation on the user's beliefs, ASE learns a model of the user's state estimation process: after each task completion, it searches for a model that would have led to beliefs that explain the user's actions.
We evaluate ASE in a user study with 12 participants who each perform four tasks: two tasks with known user biases -- bandwidth-limited image classification and a driving video game with observation delay -- and two with unknown biases that our method has to learn -- guided 2D navigation and a lunar lander teleoperation video game.
ASE's general-purpose approach to synthesizing informative observations enables a different assistance strategy to emerge in each domain, such as quickly revealing informative pixels to speed up image classification, using a dynamics model to undo observation delay in driving, identifying nearby landmarks for navigation, and exaggerating a visual indicator of tilt in the lander game.
The results show that ASE substantially improves the task performance of users with bandwidth constraints, observation delay, and other unknown biases.
\end{abstract}

\section{Introduction}

\epigraph{\itshape Chew, if only you could see what I've seen with your eyes.}{---Roy Batty, \textit{Blade Runner (1982)}}

People cannot directly access the state of the world, and must instead estimate it from sensory observations \citep{knill1996perception}.
Unfortunately, systematic biases in the user's state estimation process can lead to inaccurate beliefs and suboptimal actions.
For example, the user may not be able to keep track of many different sensors simultaneously while flying a plane \citep{mulder1999cybernetics}, or navigate with a visual impairment while listening to a smartphone guide exhaustively list all nearby objects \citep{paneels2013listen}.
Tasks performed over a network, like teleoperating space robots \citep{fong2013space}, may require the user to compensate for unintuitive, intermittent delays in observations.
Lens distortions can cause drivers to overestimate distances to obstacles: the warning, ``objects in mirror may be closer than they appear,'' is engraved into the side mirrors of cars.

Short of intervening in human cognition through brain stimulation, or training users to overcome their biases, how can we assist users with performing more accurate perception?
The key idea in this paper is that an automated assistant can intervene in human perception by modifying the observations the user receives.
Given the user's biases, different observations lead to different state estimates.
We invert this process to `trick' the person into arriving at the correct estimate: we modify observations so that, when processed by the biased user, they induce an accurate state estimate.

\begin{figure}[t]
    \centering
    \includegraphics[width=\linewidth]{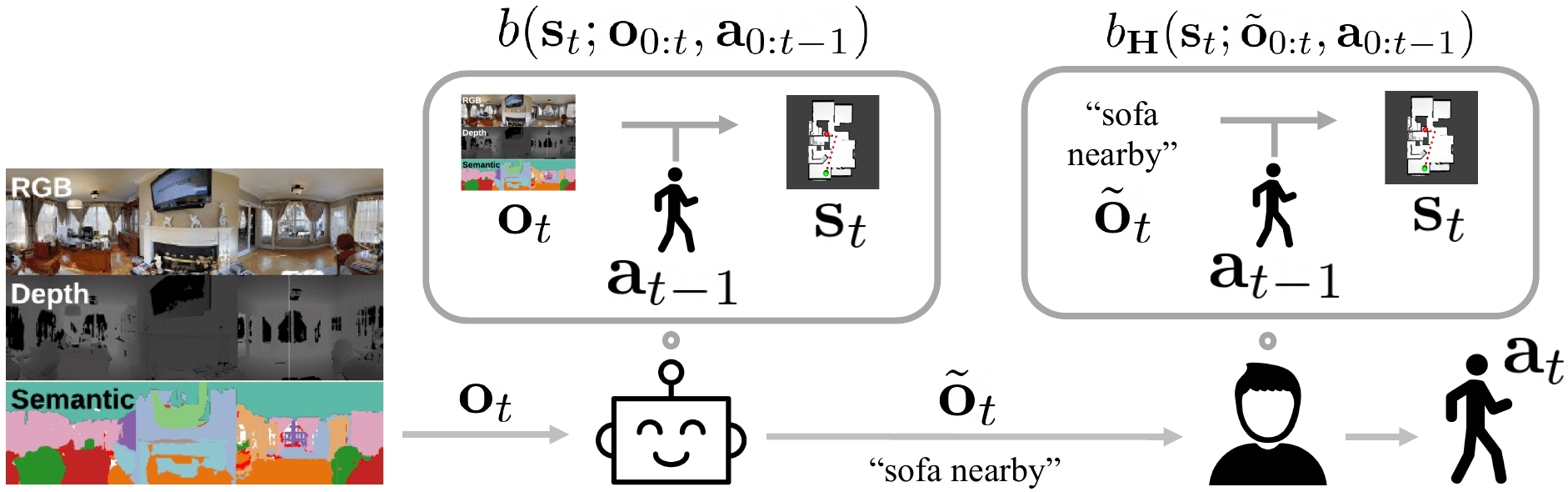}
    \caption{The assistant processes observations $\bo_t$ generated by the environment on behalf of the user \textbf{H}, updates its belief distribution over the current state $b(\bs_t; \bo_{0:t}, \ba_{0:t-1})$, then synthesizes an observation $\tilde{\bo}_t$ that will induce accurate beliefs $b_{\textbf{H}}(\bs_t; \tilde{\bo}_{0:t-1}, \tilde{\bo}_t, \ba_{0:t-1})$ when shown to the user, enabling the user to make better decisions $\ba_t$. For example, the assistant may use a smartphone camera and speaker to guide a visually-impaired user through an indoor environment: the assistant observes the user's egocentric scene through the camera, uses an object detector to determine nearby objects, then tells the user about one of them through the speaker. If the user's mental map of the environment includes object locations, the user can then infer their position and orientation: they must be in one of the states from which the mentioned object is visible. Enumerating all visible objects may overwhelm the user, so we assume the user is `bandwidth-constrained' to hearing about just one object at a time. Hence, the assistant's challenge is to select the single object that will be most informative to the user (e.g., a landmark that is only visible from the user's current state).}
    \label{fig:schematic}
\end{figure}

Figure \ref{fig:schematic} describes our method: the assistant collects observations from the environment, performs state estimation unencumbered by cognitive biases, then shows the user a synthetic, optimized observation that induces accurate beliefs when processed by their biased perception system.
These synthetic observations could be constrained to augment real observations -- for example, in an augmented reality interface \citep{zhao2019designing} -- or could completely replace them -- for instance, by replacing the user's video feed for teleoperating a robot.
Crucially, this approach does not require knowing the current task the user is performing: rather than inducing the optimal action for the task, we induce accurate state estimates, so that the user can then decide on task-appropriate actions.

The main challenge is that in order to determine how informative a synthetic observation will be to the user, we need a model of the user's state estimation process.
For instance, we might not know the user's bandwidth constraint, observation delay, or even what kinds of biases they have.
We introduce an approach for training this model online: we start assisting with an initial model, and collect data of the user suboptimally performing tasks (e.g., navigating to different goals).
We assume that upon task completion, the assistant gets to know what the task was (e.g., which goal the user was trying to reach) and can compute a near-optimal policy for that task (e.g., using reinforcement learning).
We then look in hindsight at the user's actions, and optimize model parameters that lead to state estimates which make the observed actions seem near-optimal.
Intuitively, we ask what the user must have believed, and what state estimation process would have led to those beliefs given the observations they received.
The experiments in Appendix \ref{sim-exp-BI} show that the quality of assistance improves as we collect more data and the maximum-likelihood model gets closer to the user's internal state estimation process.

Our primary contribution is the assistive state estimation (ASE) algorithm for optimizing synthetic observations to induce accurate beliefs about the current state in the user.
We evaluate ASE through a user study with 12 participants who each perform four tasks: two where the user's bias is known -- image classification from bandwidth-limited input, and a driving video game with observation delay -- and two where the bias is unknown -- navigating a 2D simulation with limited vision where the user only remembers the locations of certain objects but not others, and a lunar lander teleoperation video game.
The lander experiment is particularly interesting: the assistant learns to modify the tilt indicator away from its real value; actually improving the user's task performance, possibly because users tend to underestimate tilt.
Our user studies show that in all domains, ASE substantially improves the user's task performance, relative to a passive baseline that simply shows the user an ambient observation generated by the environment.
In addition to the user study, we perform experiments with simulated users that show ASE improves the accuracy of simulated users' internal beliefs, and that ASE is capable of learning an expressive, neural network model of the user's belief update given enough demonstration data.

\section{Assisting Users by Optimizing Observations} \label{methods}

We formulate the assistance problem as follows.
We assume that the environment follows a partially observable Markov decision process (POMDP; \citealp{kaelbling1998planning}) with state space $\mathcal{S}$, observation space $\Omega$, initial state distribution $p^{\mathrm{init}}(\bs_0)$, state transition dynamics $p^{\mathrm{dyn}}(\bs'|\bs,\ba)$, observation model $p^{\mathrm{obs}}(\bo|\bs)$, and unknown reward function $R(\bs,\ba)$.
At each timestep $t$, the assistant samples an ambient observation $\bo_t$ from the environment.
The assistant then intervenes and provides the user with a different observation $\tilde{\bo}_t \in \Omega$.
Since the reward function is unknown, we cannot compute the optimal action and provide the user with an observation that will induce them to take the optimal action.
Instead, we aim for a task-agnostic method that assists the user by providing them with an observation that efficiently communicates the current state.

Our approach to this problem is outlined in Figure \ref{fig:schematic}.
We assume that the user's state estimation process differs from the assistant's, and that this mismatch leads to suboptimal user behavior.
We assist the user by showing them synthetic observations that induce accurate beliefs about the current state.
In particular, the assistant first performs state estimation, then optimizes an observation to update the user's beliefs to match the assistant's beliefs.
To improve the assistant, we learn a personalized model of the user's state estimation process from demonstrations of suboptimal user actions on known tasks.

\subsection{Preliminaries: Assumptions about State Estimation} \label{prelims}

The standard, recursive Bayesian filter \citep{thrun2005probalistic} performs state estimation using the belief update,
\begin{equation} \label{eq:bayes-filter}
b(\bs_t|\bo_{0:t}, a_{0:t-1}) \propto p^{\mathrm{obs}}(\bo_t|\bs_t)\int_{\mathcal{S}} p^{\mathrm{dyn}}(\bs_t|\bs_{t-1},\ba_{t-1})b(\bs_{t-1} | \bo_{0:t-1}, \ba_{0:t-2})d\bs_{t-1}.
\end{equation}
In domains with a small, discrete state space $\mathcal{S}$, we compute exact belief updates using Equation \ref{eq:bayes-filter}.
In domains with high-dimensional, continuous states, the belief update in Equation \ref{eq:bayes-filter} may be intractable to compute.
To address this issue, we represent the state estimation process in continuous domains as
\begin{equation} \label{eq:continuous-belief-update}
b(\bs_t | \bo_{0:t}, \ba_{0:t-1}) = \mathcal{N}(\bs_t; \mu = f(\bo_{0:t}, \ba_{0:t-1}), \Sigma = I\sigma^2),
\end{equation}
where $f$ is a known state encoder that maps a sequence of observations and actions to a continuous, vector-valued state.
Although this procedure does not necessarily perform Bayesian belief updates, it enables us to apply our method to domains where the true initial state distribution $p^{\mathrm{init}}$, true dynamics model $p^{\mathrm{dyn}}$, and true observation model $p^{\mathrm{obs}}$ are unknown, but a state encoder $f$ is available.

\subsection{Synthesizing Observations that Induce Accurate Beliefs} \label{induce-beliefs}

To assist the user, we synthesize an observation $\tilde{\bo}_t$ such that, after the user observes $\tilde{\bo}_t$ and updates their beliefs about the current state $\bs_t$, the user's beliefs will match the assistant's beliefs.
Formally, given a history $(\bo_{0:t}, \tilde{\bo}_{0:t-1}, \ba_{0:t-1})$, the assistant decides which observation to provide to the user \textbf{H} by greedily minimizing the KL-divergence between the assistant's beliefs and the user's beliefs at the end of the current timestep:
\begin{equation} \label{eq:induce}
\tilde{\bo}_t \leftarrow \argmin_{\tilde{\bo}_t \in \Omega} \kldiv{\underbrace{b(\bs_t|\bo_{0:t}, \ba_{0:t-1})}_{\text{assistant's beliefs}}}{\underbrace{\hat{b}_{\mathbf{H}}(\bs_t|\tilde{\bo}_{0:t-1}, \tilde{\bo}_t, \ba_{0:t-1})}_{\text{assistant's prediction of user's beliefs}}},
\end{equation}
where $\hat{b}_{\mathbf{H}}$ is the assistant's model of the user's state estimation process.
The assistant's beliefs are fixed during this optimization, having already been conditioned on the most recent ambient observation $\bo_t$ generated by the environment, while the user's beliefs are conditioned on the synthetic observation $\tilde{\bo}_t$ and can thus be optimized.
The experiments in Section \ref{exp-overview} and Appendix \ref{sim-exp-BI} illustrate how different assistance strategies emerge from Equation \ref{eq:induce}, such as revealing informative pixels for image classification, undoing observation delay in driving by forward-predicting the current observation, identifying landmarks for navigation, and exaggerating indicators of dangerous states in a landing task.

\subsection{Learning Personalized Models of State Estimation} \label{learn-internal-model}

To optimize the synthetic observation in Equation \ref{eq:induce}, we need to model how the user will update their beliefs in response to observations.
We assume the user's unknown state estimation process $b_{\mathbf{H}}$ differs from the assistant's known process $b$ described in Equations \ref{eq:bayes-filter} and \ref{eq:continuous-belief-update}.
In particular, we assume $b_{\mathbf{H}}$ lies in hypothesis space $\mathcal{B}$.
The hypothesis space, which we parameterize as $\mathcal{B} = \{b_{\theta} : \theta \in \Theta\}$, captures our prior assumptions about possible user biases.
If we want to make minimal assumptions about the user's biases, we could define $\theta$ to be the weights in a neural network state encoder $f_{\theta}$ that defines the belief update $b_{\theta}$ via Equation \ref{eq:continuous-belief-update}.
If instead we assume that the user performs a Bayesian belief update on each new observation and action, but potentially ignores or misinterprets certain observations, we could define $\theta$ to be the observation probabilities $p^{\mathrm{obs}}_{\theta}(\bo | \bs) = \theta_{\bo, \bs}$ in Equation \ref{eq:bayes-filter}.
In each of our experiments, we make different assumptions about the user, leading to different choices of hypothesis space $\mathcal{B}$.

We search the hypothesis space for a model that best explains user behavior.
We assume access to a dataset $\mathcal{D}$ of demonstrations of suboptimal user actions on known tasks.
This dataset could be generated offline by the user without the assistant's help, or generated online while the assistant helps the user.
After each demonstration episode, we ask the user what task they were trying to perform during that episode.
The task could be specified, for example, through a goal state or a reward function.
Let $\tau = (\bo_{0:T-1}, \ba_{0:T-1})$ denote a demonstration, where $T$ is the episode length.
We model the user's actions as rational with respect to their beliefs about the current state:
\begin{equation} \label{eq:user-action}
p(\ba_t | \bo_{0:t}, \ba_{0:t-1}; \theta) = \int_{\mathcal{S}} \pi(\ba_t|\bs_t) b_{\theta}(\bs_t | \bo_{0:t}, \ba_{0:t-1})d\bs_t,
\end{equation}
where $\pi$ is the user's policy, which we assume to be near-optimal for their desired task.
We compute $\pi$ in hindsight after asking the user what task they were trying to demonstrate; e.g., by asking the user to write down the reward function, then doing maximum entropy reinforcement learning \citep{levine2018reinforcement}.
Note that we only need to know a near-optimal policy for the tasks in the demonstrations used to train the user model.
We do not need to know the policy at test time when we synthesize observations to assist the user.
We assume that the user's policy $\pi$ for a given task and their belief update $b_{\mathbf{H}}$ do not change once the assistant begins modifying observations using Equation \ref{eq:induce}.
In practice, even if the user adapts their policy or state estimation process to the assistant, this tends to improve performance, rather than hurt it.

We use gradient descent to compute the maximum-likelihood estimate,
\begin{equation} \label{eq:mle}
\hat{\theta} \leftarrow \argmax_{\theta} \sum_{\tau \in \mathcal{D}} \sum_t \log{p(\ba_t|\bo_{0:t}, \ba_{0:t-1}; \theta)}.
\end{equation}
We select the maximum-likelihood model to be our model of the user's state estimation process: $\hat{b}_{\mathbf{H}} \leftarrow b_{\hat{\theta}}$.
This model enables us to predict the effect of an observation on the user's beliefs.

\subsection{Assistive State Estimation}

\begin{algorithm}[t]
\begin{algorithmic}[H]
\State{Require $b_{\mathrm{init}} \in \mathcal{B}$ \Comment{initial model of user}}
\If{$\mathcal{S}$ is discrete}
\State{Require $p^{\mathrm{init}}(\bs_0), p^{\mathrm{dyn}}(\bs'|\bs,\ba), p^{\mathrm{obs}}(\bo|\bs)$ \Comment{for assistant's belief update in Equation \ref{eq:bayes-filter}}}
\ElsIf{$\mathcal{S}$ is continuous}
\State{Require state encoder $f(\bo_{0:t}, \ba_{0:t-1})$ \Comment{for assistant's belief update in Equation \ref{eq:continuous-belief-update}}}
\EndIf
\State{Initialize $\mathcal{D} \leftarrow \emptyset$ \Comment{user demonstrations}}
\State{Initialize $\hat{b}_{\mathbf{H}} \leftarrow b_{\mathrm{init}}$ \Comment{assistant's model of user}}
\While{true}
  \State{$s_0 \sim p^{\mathrm{init}}(\bs_0)$}
  \For {$t \in \{0, 1, 2, ..., T-1\}$}
    \State{$\bo_t \sim p^{\mathrm{obs}}(\bo_t|\bs_t)$ \Comment{assistant sees true observation, updates beliefs}}
    \State{$\tilde{\bo}_t \leftarrow \arg\min_{\tilde{\bo}_t \in \Omega} \kldiv{b(\bs_t | \bo_t)}{\hat{b}_{\mathbf{H}}(\bs_t | \tilde{\bo}_t)}$ \Comment{assistant synthesizes observation}}
    \State{$\ba_t \sim p(\ba_t | \tilde{\bo}_{0:t}, \ba_{0:t-1})$ \Comment{user sees synthetic observation, updates beliefs, takes action}}
    \State{$\bs_{t+1} \sim p^{\mathrm{dyn}}(\bs_{t+1}|\bs_t,\ba_t)$}
  \EndFor
  \State{$\mathcal{D} \leftarrow \mathcal{D} \cup \{(\tilde{\bo}_{0:T-1}, \ba_{0:T-1})\}$}
  \State{$\hat{\theta} \leftarrow \argmax_{\theta} \sum_{\tau \in \mathcal{D}} \sum_t \log{p(\ba_t|\tilde{\bo}_{0:t}, \ba_{0:t-1}; \theta)}$ \Comment{assistant learns model of user}}
  \State{$\hat{b}_{\mathbf{H}} \leftarrow b_{\hat{\theta}}$}
\EndWhile
\end{algorithmic}
\caption{Assistive State Estimation (ASE)}
\label{alg:ase-alg}
\end{algorithm}

Our assistive state estimation (ASE) method is summarized in Algorithm \ref{alg:ase-alg}.
We initialize the user model $\hat{b}_{\mathbf{H}}$ with an initial model $b_{\mathrm{init}}$.
In domains with a small, discrete state space $\mathcal{S}$, we assume knowledge of the initial state distribution, state transition dynamics, and observation model, in order to compute Bayesian belief updates using Equation \ref{eq:bayes-filter}.
In domains with high-dimensional, continuous states, we instead assume knowledge of a state encoder, so we can estimate the current state using Equation \ref{eq:continuous-belief-update}.
At the start of each timestep $t$, the assistant collects an observation $\bo_t$ from the environment.
The assistant then optimizes a synthetic observation $\tilde{\bo}_t$ that, when shown to the user, will induce beliefs that match the assistant's (Equation \ref{eq:induce}).
The user sees the synthetic observation $\tilde{\bo}_t$, takes an action $\ba_t$, and the environment generates the next state $\bs_{t+1}$.
At the end of each episode, we ask the user what task they were trying to perform, add the episode to the dataset $\mathcal{D}$, and re-train the user model $\hat{b}_{\mathbf{H}}$ using Equation \ref{eq:mle}.

\section{Related Work} \label{related-work}

\noindent\textbf{Modeling human beliefs, preferences, and behavior.}
Inverse planning \citep{baker2009action} and inverse reinforcement learning \citep{ng2000algorithms} learn a model of the user's reward function from demonstrated actions.
These methods typically assume that user actions are near-optimal, and can be affected by random noise \citep{ziebart2008maximum}, risk sensitivity \citep{majumdar2017risk}, or dynamics model misspecification \citep{reddy2018you}.
The closest prior work learns a reward function or policy from demonstrations, using a behavioral model that allows for false beliefs about the current state \citep{evans2016learning,schmitt2017see,daptardar2019inverse,inverseactivesensing}.
ASE differs in that we explicitly avoid trying to learn the user's task-specific reward function or policy.
Instead, we provide the user with task-agnostic assistance by learning a model of the user's state estimation process, and supplying the user with informative observations.

\noindent\textbf{Task-specific assistance via communication and visualization.}
\citet{buhler2020theory} assist users by modeling their internal beliefs and communicating observations that induce optimal actions, but require knowledge of the user's reward function at test time, assume a discrete state space, and do not learn a personalized model of the user's internal state estimation process.
ASE does not assume knowledge of the task rewards at test time, can be applied to domains with high-dimensional, continuous observations like images, and interactively learns a user model.
\citet{hilgard2019learning} learn to visualize high-dimensional examples to assist users with one-step classification tasks, whereas we focus on sequential decision-making and make minimal assumptions about the desired task.
\citet{seqinterp} use a human-in-the-loop reinforcement learning method to train an agent to sequentially explain black-box model predictions to a human auditor, where the agent is rewarded for causing the user's mental model of the predictive model to match the actual predictive model.
Our work differs in that it focuses on improving users' situational awareness in control tasks with partial observations, rather than improving model interpretability.

\noindent\textbf{Assistive navigation for visually-impaired users.}
The closest prior work plans instructional guidance actions under uncertainty about how the user will respond to instructions \citep{ohn2018personalized}.
We take a complementary approach to assistance that focuses on situational awareness: we help the user estimate their current state, so that they can make more informed decisions in general.
In particular, ASE could be useful for systems that inform users about nearby objects and points of interest through haptic or audio feedback \citep{wang2017enabling,sato2017navcog3}: as users build a mental map of their environment to support navigation \citep{banovic2013uncovering,guerreiro2017virtual}, ASE can learn a user model that captures differences between the mental map and the real environment, then prioritize information that enables the user to localize themselves and nearby obstacles, without overwhelming the user with too much information \citep{paneels2013listen}.

\section{User Studies} \label{exp-overview}

In our experiments, we evaluate whether ASE can provide helpful assistance to users; both in the case where we have prior knowledge of their state estimation process, and where we do not have such knowledge and must learn the state estimation model in the loop.
We conduct a user study with 12 participants who each perform four tasks: classifying MNIST images under bandwidth constraints \citep{lecun1998mnist}, playing the Car Racing video game from the OpenAI Gym with observation delay, navigating a simulated 2D environment with limited vision, and playing the Lunar Lander video game from the OpenAI Gym with limited vision \citep{brockman2016openai}.
We also conduct experiments with simulated users to study our method under ideal assumptions, and measure the accuracy of the simulated users' internal beliefs (Appendix \ref{sim-exps} contains details).

\subsection{Assisting Users with Known Biases} \label{exp-known}

Our first set of user studies seeks to answer \textbf{Q1}: can we assist users when we assume we know their state estimation process?
We test this hypothesis on MNIST image classification, and the Car Racing video game from the OpenAI Gym.

\subsubsection{MNIST Image Classification with a Bandwidth Constraint} \label{exp-known-mnist}

In this experiment, we test ASE's ability to assist the user when the user cannot leverage their memory of past actions and their knowledge of the state transition dynamics to infer the current state, and must rely entirely on observations.
To that end, we formulate a sequential image classification task in which the user's actions have no effect on the state.
We take the standard MNIST digit classification problem and intentionally introduce a bandwidth constraint: at each timestep, the user is shown one row of 28 pixels in the 28x28 image, and must try to classify the image given only the pixels observed so far.
The assistant observes the full image at the start of the episode, and aims to help the user classify the image as quickly as possible by showing the user informative pixels.

The assistant uses a recurrent neural network state encoder $f$ to compute the belief update $b$ via Equation \ref{eq:continuous-belief-update}, where $f$ is trained offline to reconstruct the full image given a sequence of pixel observations.
We assume that the user's belief update $b_{\mathbf{H}}$ is equivalent to the assistant's belief update $b$, except that it can only process one row of pixels per timestep.
We compute the optimal synthetic observation $\tilde{\bo}_t$ by simply enumerating all rows of pixels that have not been shown to the user yet, and computing the KL-divergence (Equation \ref{eq:induce}) for each possible value of $\tilde{\bo}_t$.
Appendix \ref{imp-deets} describes the experimental setup in more detail.

\noindent\textbf{Manipulated factors.}
We evaluate (1) an unassisted baseline that reveals the pixel rows in order from top to bottom; (2) a random baseline that reveals a new pixel row sampled uniformly at random; and (3) ASE.

\noindent\textbf{Dependent measures.}
We measure the user's classification accuracy at each timestep, in order to capture how quickly the user recognizes the image over the course of an episode.

\noindent\textbf{Subject allocation.}
We recruited 11 male and 1 female participants, with an average age of 25.
Each participant was provided with the rules of the task and example images, then labeled 25 different digits.
Each digit was broken down into an episode of 28 partial images, yielding a total of 700 labels per user.
To avoid the confounding effect of users learning to classify images more accurately and quickly over time, we randomly interleave episodes from each of the three conditions.
For example, episode 1 is unassisted, episode 2 is assisted by ASE, episode 3 is assisted by the random baseline, etc.

\begin{figure}[t]
  \centering
  \includegraphics[width=\linewidth]{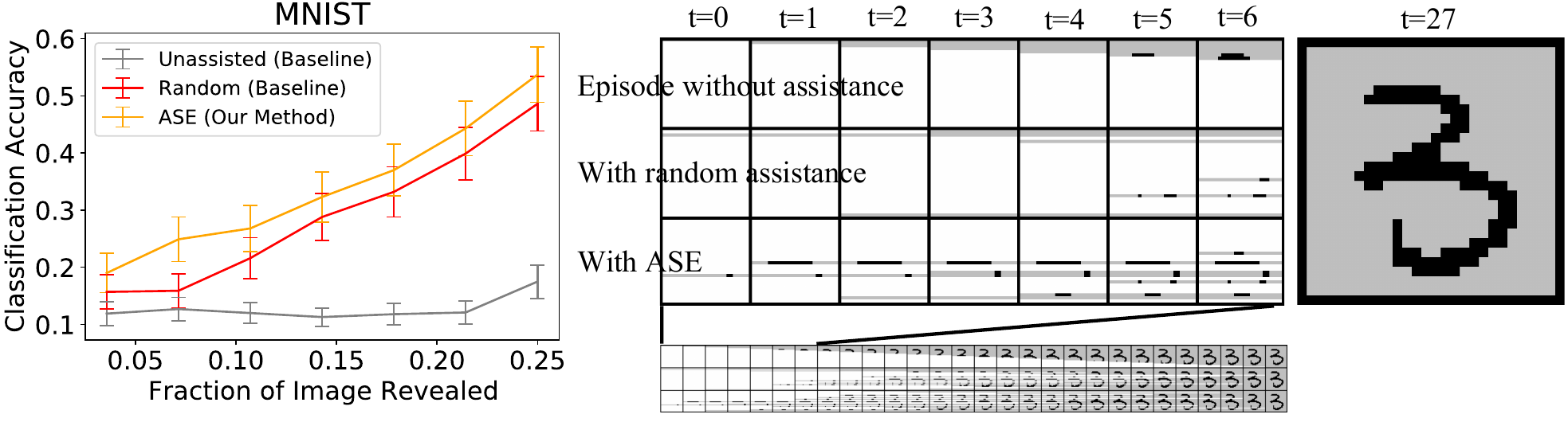}
  \caption{MNIST image classification experiments that address \textbf{Q1} -- can we assist users when we assume we know their state estimation process? -- by comparing our method (ASE), which synthesizes an informative observation under the assumption that the user's belief update is similar to the assistant's, to baselines that either use ambient observations generated by the environment (Unassisted) or randomly generate observations (Random). ASE tends to quickly reveal rows near the middle and rows with many non-zero pixels, enabling the user to more accurately guess the label earlier. In the unassisted condition, revealing rows in order from top to bottom is not as quick to reveal informative pixels. The random baseline tends to spread them out uniformly throughout the image, which is a good strategy in the long run but does not necessarily reveal informative pixels early in the episode. We measure standard error across 100 episodes.}
  \label{fig:mnist-results}
\end{figure}

\noindent\textbf{Analysis.}
Figure \ref{fig:mnist-results} shows that ASE substantially outperforms the unassisted baseline (orange vs. gray curve), and enables the user to classify the digit using fewer timesteps (i.e., fewer pixels) than the random baseline (orange vs. red curve).
We ran a one-way repeated measures ANOVA on the classification accuracy dependent measure from the random and ASE conditions, with the presence of ASE as a factor and the digit ID and fraction of image revealed as covariates, and found that $f(1, 5452) = 7.97, p < .01$.
While the effect was not substantial -- the assisted user's least-squares-mean accuracy was 74.2\%, while the unassisted user's was 71.7\% -- the assisted user achieved significantly higher accuracy than the unassisted user.
Although the uniform-random baseline happens to perform well on MNIST, it performs extremely poorly in simulation experiments with 2D navigation and Car Racing (Table \ref{tab:hab-sim-results} and Figure \ref{fig:car-sim-results} in the appendix).

\subsubsection{Car Racing Video Game with Observation Delay}

\begin{figure}[t]
  \centering
  \includegraphics[width=\linewidth]{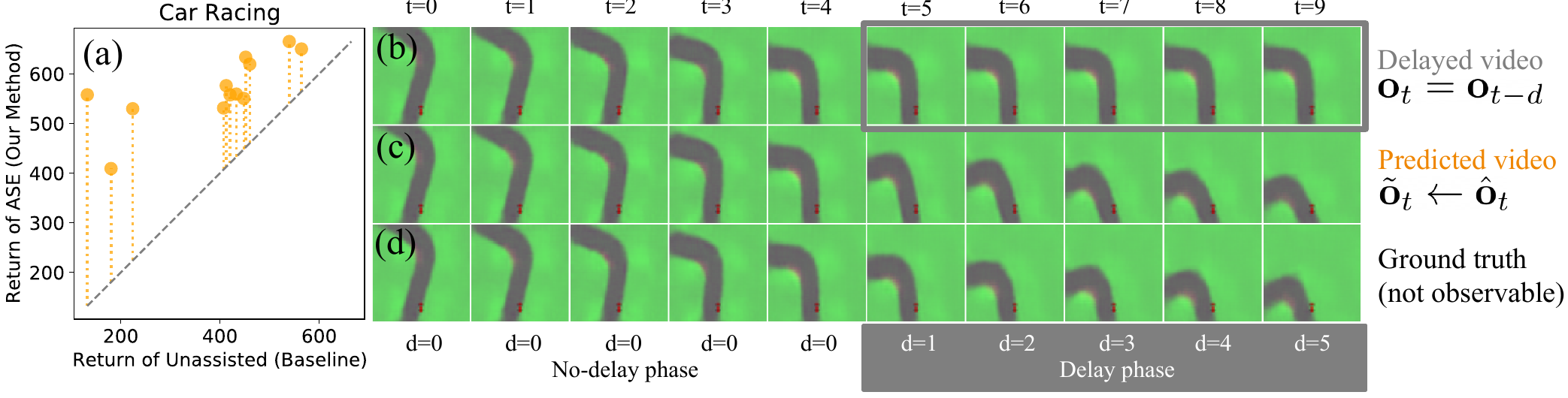}
  \caption{Car Racing video game experiments that address \textbf{Q1} -- can we assist users when we assume we know their state estimation process? -- by comparing our method (ASE), which synthesizes an informative observation under the assumption that the user's belief update is similar to the assistant's, to a baseline that always shows the ambient observation generated by the environment (Unassisted). \textbf{(a)} Each orange circle represents one of the 12 participants. The dashed gray line shows baseline-equivalent performance, and the dotted orange lines show the difference between assisted and unassisted performance. Per-user return is averaged across 3 episodes (50 seconds each). \textbf{(b-d)} Top-down views of approaching a left turn with observation delay $d$ at time $t$: \textbf{(b)} outdated ambient observation $\bo_t$, \textbf{(c)} forward-predicted observation representative of the current state $\hat{\bo}_t$, and \textbf{(d)} the ground truth, which cannot be observed by either the user or the assistant. ASE shows the user the forward-predicted observation $\hat{\bo}_t$, which is closer to the ground truth than the outdated ambient observation $\bo_t$ that the user would see by default, especially when the delay is $d$ is large.}
  \label{fig:car-results}
\end{figure}

In this experiment, we test ASE's ability to assist the user in a real-time driving game with delayed observations, where the user tends to react to outdated observations as if they are current.
Our assistant sees the same delayed observations as the user, but instead of passing them to the user, replaces the user's video feed with synthetic images produced by a generative model.
To optimize these images to induce the correct state beliefs in the user, the assistant forward-predicts the current state from the delayed observation and the user's most recent actions, then constructs an image observation representative of the predicted current state.
By default, this environment emits a 64x64 RGB image observation with a top-down view of the car, and the user can steer left or right using their keyboard (Figure \ref{fig:car-results}).
To simulate intermittent observation delays, we set up the environment to alternate between a no-delay phase of emitting new observations immediately (for 5 timesteps) and a delay phase of repeatedly emitting the final observation from the previous no-delay phase (for 5 timesteps).
Both the assistant and the user experience the same delay.

The assistant uses a recurrent neural network (RNN) state encoder $f$ to compute the belief update $b$ via Equation \ref{eq:continuous-belief-update}, and a variational auto-encoder (VAE; \citealp{kingma2013auto}) to synthesize image observations from the hidden states of $f$.
We assume that the user's belief update $b_{\mathbf{H}}$ is identical to the assistant's belief update $b$, except that $b_{\mathbf{H}}$ treats observations as if they are never delayed.
In practice, although users can clearly tell there is a delay, they are incapable of adjusting to it and steer as if there is no delay.
If the last $d$ observations are delayed, a straightforward solution emerges from the assistant's belief-matching objective in Equation \ref{eq:induce}: replace the delayed observations $\bo_{t-d+1:t}$ with recursively predicted, non-delayed observations $\hat{\bo}_{t-d+1:t}$ from the RNN encoder $f$ and VAE image decoder, and show the user the prediction of the current observation: $\tilde{\bo}_t \leftarrow \hat{\bo}_t$.
If the last observation $\bo_t$ was not delayed, then the assistant simply shows the ambient observation: $\tilde{\bo}_t \leftarrow \bo_t$.
Appendix \ref{imp-deets} describes the experimental setup in more detail.

\noindent\textbf{Manipulated factors.}
We evaluate (1) an unassisted baseline that passively shows the ambient observation generated by the environment and (2) ASE.

\noindent\textbf{Dependent measures.}
We measure performance using a reward function that penalizes going off road and gives bonuses for visiting new patches of the road.

\noindent\textbf{Subject allocation.}
We recruited 11 male and 1 female participants, with an average age of 25.
Each participant was provided with the rules of the task and a short practice period of 2 episodes to familiarize themselves with the controls and dynamics.
Each user played in both conditions: unassisted, and assisted by ASE.
To avoid the confounding effect of users learning to play the game better over time, we counterbalanced the order of the two conditions.
Each condition lasted 3 episodes, with 1000 timesteps (50 seconds) per episode.

\noindent\textbf{Analysis}
Plot (a) in Figure \ref{fig:car-results} shows that users are able to achieve substantially larger returns (i.e., drive on the road and stay off the grass more often) with the ASE assistant compared to the unassisted condition.
ASE makes the user's video feed smoother by predicting the current observation when the true current observation is delayed, which makes real-time, closed-loop control of the car substantially easier.
Users in the unassisted condition tended not to change their steering action when the true images were delayed, while assisted users were able to rapidly switch steering actions even during delay phases, by responding to the assistant's synthetic images.
We ran a one-way repeated measures ANOVA on the returns from the unassisted and ASE conditions with the presence of ASE as a factor, and found that $f(1, 11) = 41.01, p < .001$.
The assisted user achieved significantly higher returns than the unassisted user.
The subjective evaluations in Table \ref{tab:car-survey} in the appendix corroborate these results: users reported perceiving smaller delays and feeling more in control of the car when they were assisted.
One reason that users may have perceived a small delay even in the assisted condition is that the assistant uses an imperfect, learned state encoder $f$ in its belief update.
This suggests that even when the assistant has an imperfect state estimation process, ASE can still improve the user's task performance; the assistant's process just has to be more accurate than the user's.

\subsection{Learning to Assist Users with Unknown Biases} \label{exp-unknown}

Our second set of user studies seeks to answer \textbf{Q2}: can we assist users when we do not know their state estimation process, and must learn a model of it?
We test this hypothesis first in a 2D navigation task, then in a variant of the Lunar Lander video game from the OpenAI Gym.

\subsubsection{2D Navigation with Incomplete Mental Map of Object Locations}

\begin{figure}[t]
    \centering
    \includegraphics[width=\linewidth]{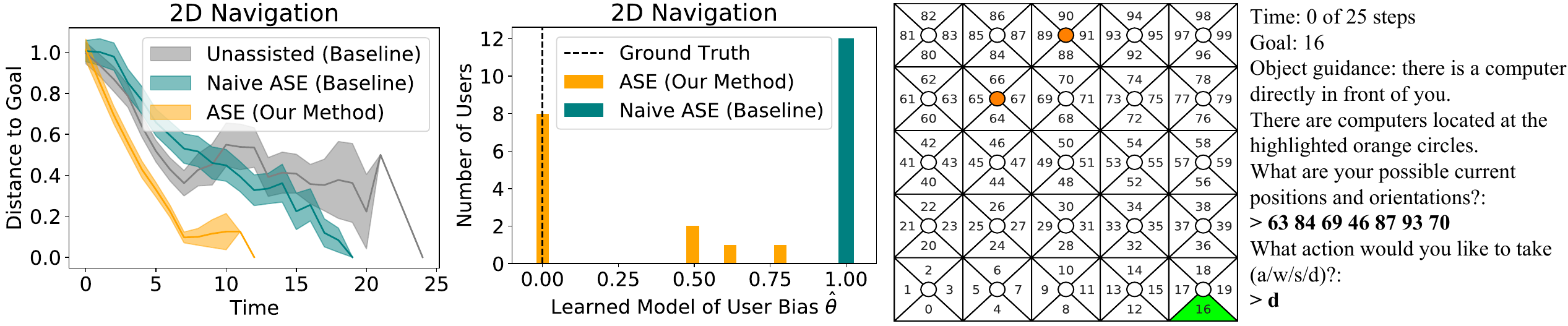}
  \caption{2D navigation experiments that address \textbf{Q2} -- can we assist users when we do not know their state estimation process, and must learn a model of it? -- by comparing our method (ASE), which learns a model of the user's belief update then synthesizes observations that are informative under the learned model, to baselines that either use ambient observations generated by the environment (Unassisted) or assume the user's belief update is similar to the assistant's and do not learn a model (Na\"{i}ve ASE). We measure standard error across 55 episodes (5 episodes per user). The results show that ASE is able to learn the user's bias parameter $\theta$, which enables the personalized assistant to give the user more informative observations than the naive assistant. The user's console interface shows them the goal state (green) and the locations of the currently-observed object in their mental map (orange circles). The user can choose to move forward (w), turn 90 degrees (a/d), or stay still and wait for another observation (s).}
    \label{fig:gw-results}
\end{figure}

In this experiment, we intentionally introduce a bias into the user's perception (unknown to ASE), and test whether ASE can learn a user model that recovers this bias.
Inspired by the assistive navigation systems discussed in Section \ref{related-work}, which inform visually-impaired users about nearby points of interest through audio feedback, we set up a simulated 2D navigation task in which the user cannot directly access their current position and orientation, but can infer them using text observations that describe nearby objects.
To incept a controlled user bias, we intentionally do not include the locations of certain objects in the user's `mental map,' which prevents the user from using observations of those objects to infer their current state as they navigate to a goal.
To effectively assist the user, ASE must learn that the user ignores observations that mention these unknown objects.
Figure \ref{fig:gw-results} illustrates the `mental map' of the 5x5 grid world shown to the user.
At each timestep, the user is told about one of the objects directly in front of them.
Some objects are unique, while other objects have multiple instances that exist in different locations (e.g., one computer vs. multiple plants).
The objects are divided into 3 categories: (a) unique but unknown, (b) not unique but known, and (c) both unique and known.

The assistant knows the locations of all objects, and can observe all objects in front of the user simultaneously.
Following Section \ref{learn-internal-model}, we parameterize the user model $b_{\theta}$ as a Bayesian belief update (Equation \ref{eq:bayes-filter}) that uses observation model $p^{\mathrm{obs}}_{\theta}(\bo | \bs)$.
The parameter $\theta \in [0, 1]$ weights the observation probabilities of objects in category (a): $p^{\mathrm{obs}}_{\theta}(\bo | \bs) \propto \theta \cdot p^{\mathrm{obs}}(\bo | \bs)$ for all objects $\bo$ in category (a).
Because we intentionally make category (a) objects unknown to the user, we know the true value: $\theta = 0$.
We would like ASE to learn this value from the user's behavior.
Appendix \ref{imp-deets} describes the experimental setup in more detail.

\noindent\textbf{Manipulated factors.}
We evaluate (1) an unassisted baseline that passively shows the ambient observation generated by the environment; (2) a na\"{i}ve version of ASE that does not train the user model, and instead continues using the initial model $b_{\mathrm{init}}$ where $\theta = 1$; and (3) ASE.
In ASE, we learn $\theta$ from the episodes collected in conditions 1 and 2.
In practice, we pool the data from the first $k$ participants to train the model for the $k$-th participant, since the small amount of data collected for each individual user is too noisy to learn an accurate model from, and because the true model does not vary between users.

\noindent\textbf{Dependent measures.}
We measure the distance from the user's current position to their goal position (normalized by distance from initial position to goal position) at each timestep, in order to capture how quickly the user moves toward the goal throughout the episode.

\noindent\textbf{Subject allocation.}
We recruited 11 male and 1 female participants, with an average age of 25.
Each participant was provided with the rules of the task and a short practice period of 3 episodes to familiarize themselves with the controls and dynamics.
Each user played in all three conditions: unassisted, assisted by na\"{i}ve ASE, and assisted by ASE.
We counterbalanced the order of the unassisted and na\"{i}ve ASE conditions.
We could not counterbalance the order of the ASE condition to control for the learning effect, since ASE learns $\hat{\theta}$ from the data collected in the unassisted and na\"{i}ve ASE conditions.
Figure \ref{fig:learning-effect} in the appendix shows that the introduction of the ASE assistant sharply improves the user's performance across episodes, suggesting the learning effect was not a substantial confounder.
Each condition lasted 5 episodes, with 25 timesteps per episode.

\noindent\textbf{Analysis.}
Figure \ref{fig:gw-results} shows that users are able to move toward the goal substantially faster with the ASE assistant compared to the unassisted condition.
Furthermore, learning a model of the user's observation model substantially improved the assistant's performance compared to the na\"{i}ve assistant.
In the unassisted condition, the user receives many observations of objects in category (b), which are known but relatively uninformative since they have multiple known locations.
In the na\"{i}ve condition, the user receives many observations of objects in categories (a), which are unique but unknown.
ASE learns that objects in category (a) are unknown (i.e., $\hat{\theta} = 0$), so it only shows the unique and known objects in category (c).
We ran a one-way repeated measures ANOVA on the time-to-goal dependent measure from the unassisted and ASE conditions with the presence of ASE as a factor, and found that $f(1, 11) = 18.02, p < .01$.
The assisted user reached the goal significantly faster than the unassisted user.
The subjective evaluations in Table \ref{tab:gw-survey} in the appendix corroborate these results: users reported finding the observations more helpful in the ASE condition compared to the unassisted condition.

\subsubsection{Lunar Lander Video Game with Limited Vision}

\begin{figure}[t]
\centering
\includegraphics[width=\textwidth]{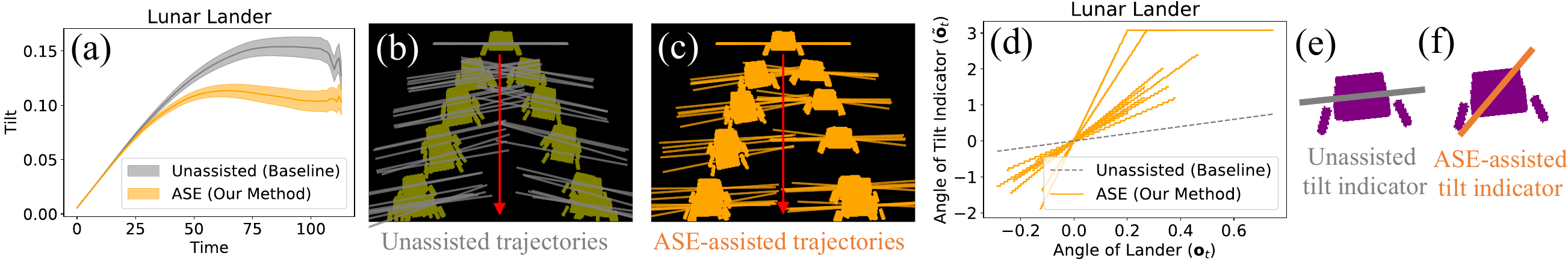}
\caption{Lunar Lander experiments that address \textbf{Q2} -- can we assist users when we do not know their state estimation process, and must learn a model of it? -- by comparing our method (ASE), which learns a model of the user's belief update then synthesizes observations that are informative under the learned model, to a baseline that always shows the ambient observation generated by the environment (Unassisted). \textbf{(a)} We measure standard error across 120 episodes (10 episodes per user). \textbf{(b)} Sample of unassisted trajectories from the user studies. \textbf{(c)} With assistance, the user keeps the lander more level. \textbf{(d-f)} ASE tends to exaggerate the tilt indicator (orange vs. gray line), and personalizes the exaggeration to the user (each orange line in \textbf{(d)} corresponds to a different user).}
\label{fig:lander-results}
\end{figure}

In this experiment, we evaluate whether ASE can learn a personalized model of naturally-occurring user biases in the Lunar Lander game, in which users tend to land at an unsafe angle.
We conjecture that this suboptimal user behavior is caused by underestimating the lander's tilt, and that the assistant might learn to help the user by showing them an image in which the lander's tilt is exaggerated beyond the ground truth.
At each timestep, the environment emits an image of the lander, and the user can fire the left or right thruster using their keyboard (plot (b) in Figure \ref{fig:lander-results}).
The objective is to make sure the lander stays level as it descends, using the thrusters to prevent it from tilting left or right.
The image includes a visual indicator of the lander's tilt, which is separate from the body of the lander (plot (e) in Figure \ref{fig:lander-results}).
The assistant is capable of freely changing the angle of this tilt indicator in the image observation shown to the user, but cannot change any other aspect of the image (e.g., the lander body itself).

To simplify our model of the user, we focus on one feature: the angle of the lander.
In the user model, an observation is characterized by the angle of the tilt indicator: the observation space is $\Omega = [-\pi, \pi]$.
A state is characterized by the lander's angle: the state space is $\mathcal{S} = [-\pi, \pi]$.
By default, the angle of the tilt indicator is equal to the lander's angle: $p^{\mathrm{obs}}(\bo|\bs) = \mathbbm{1}[\bo = \bs]$.
We assume the user's suboptimality stems from incorrectly inferring the lander's tilt from the angle of the tilt indicator: it is easy to tell when the lander is severely tilted, but harder to tell when the lander is only slightly tilted.
This is a problem for the user, since keeping the lander level requires detecting tilt early when it is still small, so that the thrusters have enough time to force the lander upright.
We represent the user model $b_{\theta}$ using a simple logistic model: $b_{\theta}(\bo_{0:t}, \ba_{0:t-1}) = -\pi + 2\pi \cdot \sigma(\theta_0 + \theta_1 \cdot \bo_t)$, where $\sigma$ is the sigmoid function.
The optimal synthetic observation anticipates the user's internal distortion: $\tilde{\bo}_t \leftarrow b^{-1}_{\hat{\theta}}(\bo_t)$.
Appendix \ref{imp-deets} describes the experimental setup in more detail.

\noindent\textbf{Manipulated factors.}
We evaluate (1) an unassisted baseline that rotates the tilt indicator to exactly match the lander's angle and (2) ASE.
In ASE, we learn $\theta$ from the episodes collected in the unassisted condition, as well as 5 assisted episodes generated iteratively using Algorithm \ref{alg:ase-alg}.

\noindent\textbf{Dependent measures.}
We measure the absolute value of the lander's angle $|\bs_t|$ (i.e., `tilt') at each timestep, in order to capture how well the user is able to stabilize the lander throughout the episode.

\noindent\textbf{Subject allocation.}
We recruited 11 male and 1 female participants, with an average age of 25.
Each participant was provided with the rules of the task and a short practice period of 5 episodes to familiarize themselves with the controls and dynamics.
Each user played in both conditions: unassisted, then assisted by ASE.
We could not counterbalance the order of the two conditions to control for the learning effect, since ASE learns $\hat{\theta}$ from the data collected in the unassisted condition.
Figure \ref{fig:learning-effect} in the appendix shows that the introduction of the ASE assistant sharply improves the user's performance across episodes, suggesting the learning effect was not a substantial confounder.
Each condition lasted 10 episodes, with 150 timesteps (10 seconds) per episode.

\noindent\textbf{Analysis.}
Plot (a) in Figure \ref{fig:lander-results} shows that users are able to substantially decrease the tilt
of the lander throughout the episode with the ASE assistant compared to the unassisted condition.
The assistant learns that the user infers a smaller lander angle than the observed tilt indicator's angle.
This leads to an assistance policy that exaggerates observations by rotating the tilt indicator to exceed the lander's true angle.
Furthermore, plot (d) in Figure \ref{fig:lander-results} shows that the learned distortion model varies across users.
We ran a one-way repeated measures ANOVA on the average tilt dependent measure from the unassisted and ASE conditions with the presence of ASE as a factor, and found that $f(1, 11) = 6.30, p < .05$.
The assisted user's average tilt was significantly smaller than the unassisted user's.
The subjective evaluations in Table \ref{tab:lander-survey} in the appendix corroborate these results: users reported finding it easier to tell when the lander was tilted in the ASE condition compared to the unassisted condition.

\section{Discussion}

\noindent\textbf{Summary.}
We propose the assistive state estimation (ASE) algorithm for helping users with perception in partially observable Markov decision processes.
The key idea is to synthesize observations that induce more accurate beliefs about the current state than the ambient observations.
In our first set of user studies, we show that ASE can assist users with limited sensor bandwidth by identifying subsets of informative pixels for image classification, and that ASE can assist users with observation delay in a driving video game by using a dynamics model to predict the current state of the world and constructing a hypothetical current observation.
In our second set of user studies, we show that ASE can assist irrational users with navigating a 2D world by informing them about nearby landmarks, and detecting and minimizing tilt in the Lunar Lander game by exaggerating a visual indicator of the lander's tilt.
These experiments broadly illustrate how assisting users with state estimation while making minimal assumptions about the desired task can improve real users' task performance.
In addition to the user studies, we run simulation experiments on indoor navigation and MNIST digit classification that show (1) ASE not only improves users' task performance, but also improves the accuracy of simulated users' internal beliefs; and (2) the quality of assistance increases with the number of user-in-the-loop episodes collected.

\noindent\textbf{Limitations and future work.}
ASE assumes that we can solve for a near-optimal state estimator in order to compute the assistant's belief update $b$ via Equations \ref{eq:bayes-filter} or \ref{eq:continuous-belief-update}, and can solve for near-optimal policies $\pi$ in order to model user actions via Equation \ref{eq:user-action}, which may not be feasible in real-world domains.
Fortunately, recent work has demonstrated substantial improvements in learning approximate state estimators and policies in complex environments with image observations, unknown dynamics, and other challenges \citep{zhang2019solar,hafner2019dream}.
While our proof of concept does not make use of these advances, incorporating them into a more practical assistive state estimation system is a promising direction for future work.
Furthermore, our user studies are limited in that we do not know if the improvement in users' task performance is caused by more accurate internal beliefs, or by some other feature of the assistance condition, since we cannot directly measure those beliefs in real users (only in simulated users).
Our ultimate goal is for the user to make better decisions when assisted, and the results show that the belief-matching objective in Equation \ref{eq:induce} accomplishes this in four different domains.
Even so, one direction for future work is to design experiments that more directly measure the user's internal beliefs during decision-making.

\section{Acknowledgements}

Thanks to members of the InterACT and RAIL labs at UC Berkeley for feedback on this project.
This work was supported in part by an NVIDIA Graduate Fellowship, NSF IIS-1700696, AFOSR FA9550-17-1-0308, and NSF NRI 1734633.

\bibliography{master}

\clearpage

\appendix

\section{Appendix}

\subsection{Simulation Experiments} \label{sim-exps}

One of the drawbacks of running a user study with human participants is that, while we can measure task performance, we cannot directly measure the accuracy of users' internal beliefs about the current state.
Studying how ASE scales with the amount of observation delay, training data, and other factors would also require a prohibitive number of human-in-the-loop experiments.
To that end, we run experiments with simulated users on indoor navigation and MNIST digit classification.

\subsubsection{Improving Accuracy of Users' Internal Beliefs} \label{sim-exp-known}

Our third experiment seeks to answer \textbf{Q3}: can we improve the accuracy of simulated users' internal beliefs?
We test this hypothesis in an indoor navigation task with a more realistic environment than the 5x5 layout from the user study: we take one floor of a 3D house from the Matterport3D dataset \citep{Matterport3D}, and discretize it into a navigable 2D grid using the Habitat framework \citep{habitat19iccv}.
We simulate the user using a goal-conditioned policy $\pi(a|s;g) \propto \exp{(Q(s,a;g))}$ that takes the shortest path to the goal, where the value function $Q$ is computed using tabular soft Q-iteration \citep{watkins1992q,bloem2014infinite} with a reward function that gives a constant negative penalty for each state transition that does not reach the goal.
The simulated user's belief update $b_{\mathbf{H}}$ is identical to the assistant's belief update $b$, except that it ignores any observation that consists of more than one object.
The assistant knows the simulated user's belief update $b_{\mathbf{H}}$.
Note that this differs from the 5x5 experiment in Section \ref{exp-unknown}, in which the user's belief update was not only bandwidth-constrained, but also tainted by a misspecified observation model due to the presence of unknown objects whose locations were not plotted in the user's mental map.
The purpose of this experiment is not to test if ASE can learn a user model, but rather to test the accuracy of the user's induced beliefs.
Appendix \ref{imp-deets} describes the experimental setup in more detail.

\noindent\textbf{Manipulated factors.}
We evaluate (1) an unassisted baseline that passively shows the ambient observation generated by the environment and (2) ASE.

\noindent\textbf{Dependent measures.}
We measure success rate, distance to the goal at the end of the episode, number of steps taken to reach the goal, and the user's internal log-likelihood of the true state.

\begin{figure}[t]
    \centering
    \includegraphics[width=\linewidth]{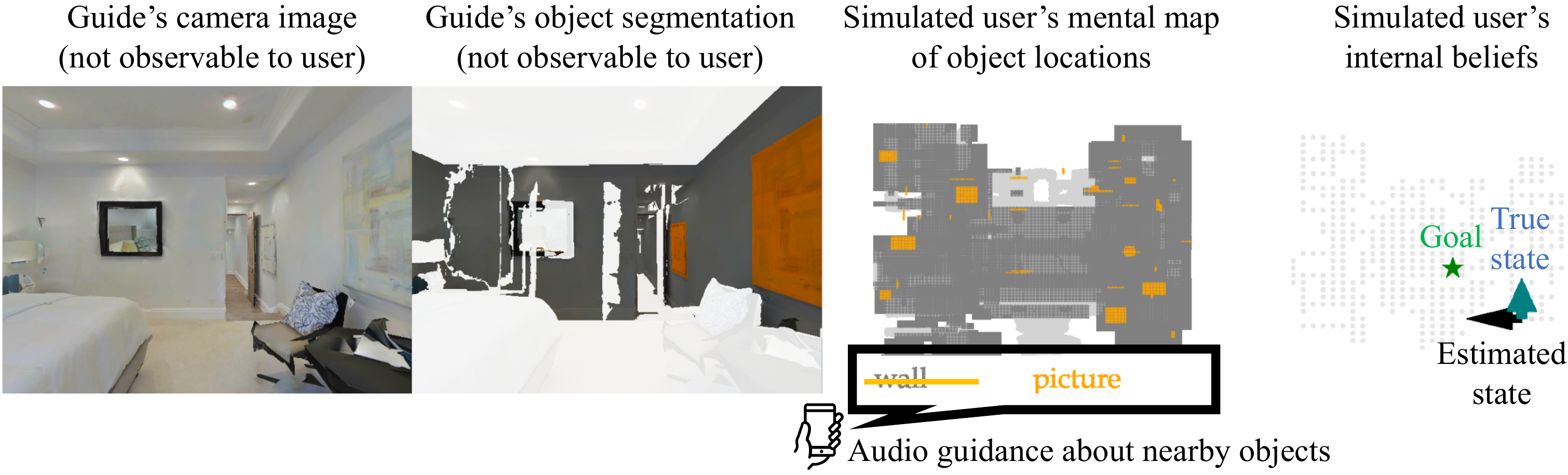}
    \caption{A simulated blind user navigating an indoor environment, using audio guidance about nearby objects to estimate position and orientation (the same problem setting as Figure \ref{fig:schematic}). The assistant sees the RGB camera image, uses the semantic mesh from the dataset to determine the list of visible objects, then replaces the ambient observation (gray), which was sampled uniformly at random from the list of visible objects, with an optimized observation (orange) that minimizes KL-divergence (Equation \ref{eq:induce}). The simulated user knows the locations of all objects, and can use this mental map to infer their current position and orientation given observations of nearby objects and memory of past movements.}
    \label{fig:hab-viz}
\end{figure}

\begin{table}[t]
    \centering
    \small
    \begin{tabular}{llllll}
    \toprule
    & Success Rate & Distance to Goal & Time to Goal & Belief in True State \\
    \midrule
Unassisted (Baseline) & $0.73 \pm 0.04$ & $0.10 \pm 0.02$ & $70.87 \pm 2.72$ & $-1.70 \pm 0.02$ \\
Random (Baseline) & $0.02 \pm 0.01$ & $1.00 \pm 0.04$ & $99.99 \pm 0.71$ & $-20.44 \pm 0.08$ \\
\rowcolor{orange} ASE (Our Method) & $1.00 \pm 0.00$ & $0.00 \pm 0.00$ & $38.55 \pm 1.60$ & $-0.72 \pm 0.02$ \\
    \bottomrule
  \end{tabular}
  \normalsize
    \caption{Habitat navigation experiments that address \textbf{Q3} -- can we improve the accuracy of simulated users' internal beliefs? -- by comparing our method (ASE), which synthesizes an informative observation that fits within the simulated user's sensor bandwidth, to baselines that either use ambient observations generated by the environment (Unassisted) or randomly generate observations (Random). The results show that our method (ASE) substantially outperforms the baselines (Unassisted and Random). The simulated user's internal beliefs are represented as log-likelihoods. We measure standard error across 100 evaluation episodes.}
    \label{tab:hab-sim-results}
\end{table}

\noindent\textbf{Analysis.}
Table \ref{tab:hab-sim-results} show that ASE substantially outperforms the unassisted and random baselines in improving the accuracy of the user's internal beliefs.
ASE tends to inform the user of landmark objects that are more likely to be seen from the current state than in other states -- like gym equipment, paintings, and showers -- enabling the user to infer the current state more accurately.
In the unassisted condition, the user tends to receive observations of objects that are common not only in the current state but also common across states -- like walls, floors, and ceilings -- which makes it difficult to identify the current state.
This result illustrates that ASE can be used to improve situational awareness, independent of the user's desired task.

\subsubsection{Scaling to Multivariate User Models} \label{sim-exp-BI}

Our fourth experiment seeks to answer \textbf{Q4}: given enough demonstration data, can ASE learn complex models of the user's state estimation process?
We test this hypothesis in the MNIST domain from Section \ref{exp-known}.
We simulate a user by training an LSTM sequence model to predict the image label given a sequence of pixel observations.
We define the simulated user's belief update $b_{\mathbf{H}}$ using Equation \ref{eq:continuous-belief-update}, where the state encoder $f_{\mathbf{H}}$ maps a sequence of observations to a 32-dimensional hidden state.
This hidden state is distinct from the hidden state produced by the assistant's state encoder $f$, due to the difference between the assistant's reconstruction objective (described in Section \ref{exp-known}) and the simulated user's classification objective.
ASE represents the user's state encoder as a recurrent neural network $f_{\theta}$ with 32 hidden units, and defines the user model $b_{\theta}$ via Equation \ref{eq:continuous-belief-update}.
Appendix \ref{imp-deets} describes the experimental setup in more detail.

\noindent\textbf{Manipulated factors.}
We evaluate (1) an unassisted baseline that passively shows the ambient observation generated by the environment; (2) a na\"{i}ve version of ASE that does not train the user model; and (3) ASE, where we learn $\theta$ from episodes generated iteratively using Algorithm \ref{alg:ase-alg}.
The na\"{i}ve assistant incorrectly assumes the user's state encoder $f_{\mathbf{H}}$ is equivalent to the assistant's state encoder $f$, except that it can only process one row of pixels per timestep.
We vary the number of training episodes $|\mathcal{D}|$ in the ASE condition.

\noindent\textbf{Dependent measures.}
We measure per-timestep classification accuracy, as in Section \ref{exp-known-mnist}.

\begin{figure}[t]
    \centering
    \includegraphics[width=0.49\linewidth]{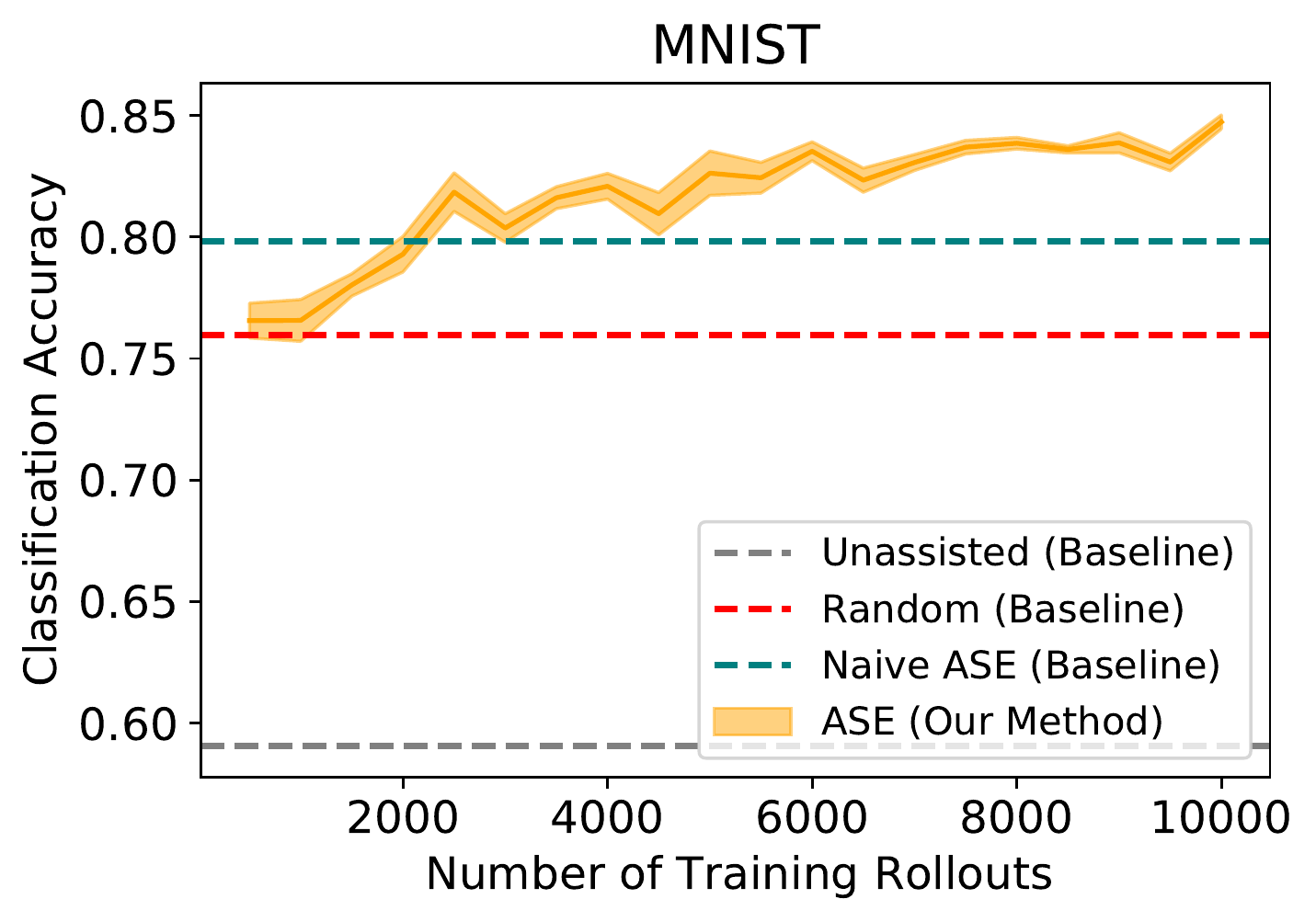}
    \includegraphics[width=0.49\linewidth]{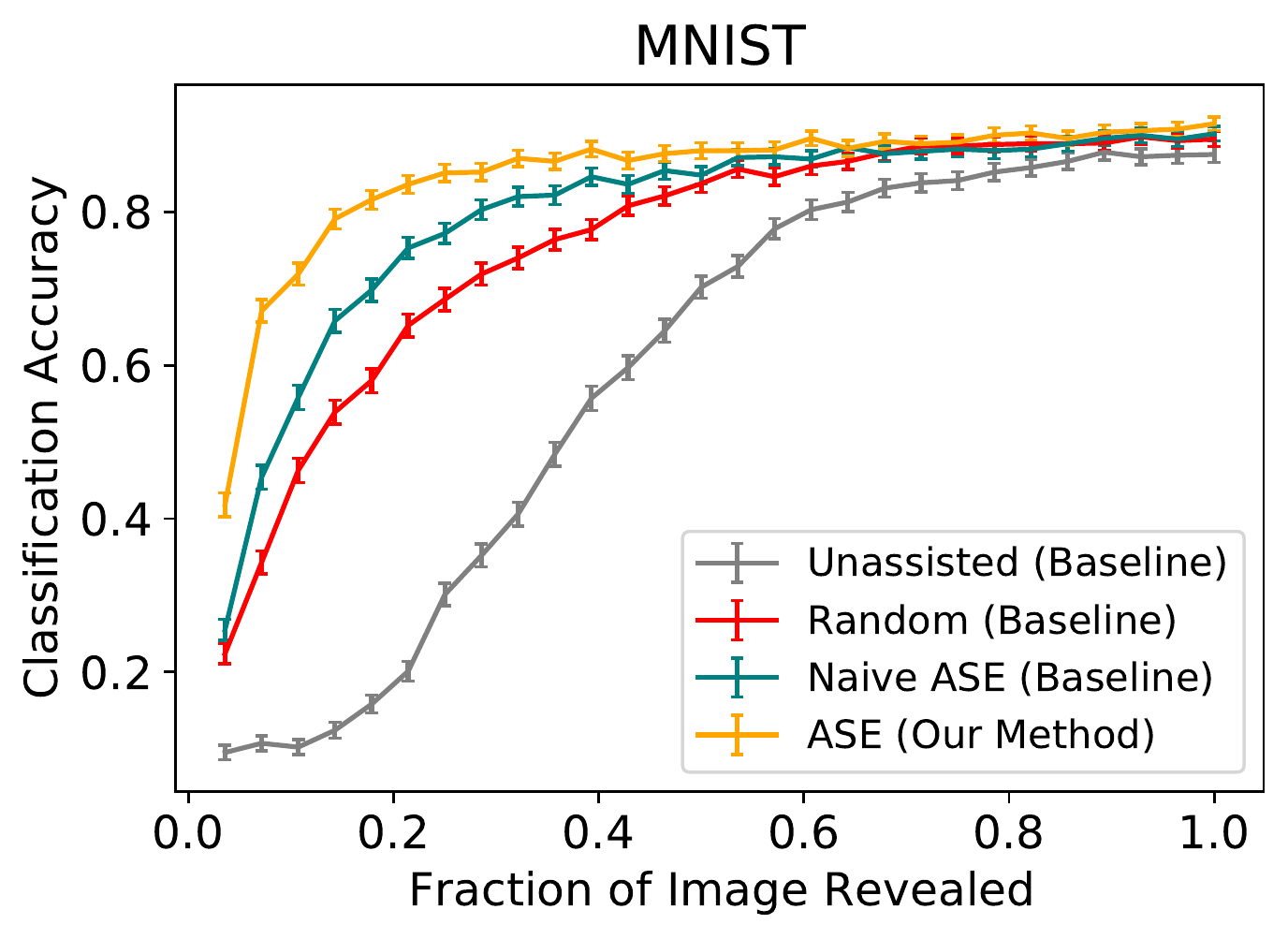}
    \caption{MNIST experiments that address \textbf{Q4} -- given enough demonstration data, can ASE learn complex models of the user's state estimation process? -- by comparing our method (ASE), which learns a model of the simulated user, to a baseline variant of our method that does not learn a model (Na\"{i}ve ASE). The results show that with enough training data, the personalized assistant outperforms the na\"{i}ve assistant by more accurately predicting the effect of a given observation on the simulated user, and thus providing more informative observations to the simulated user. We measure standard error across 5 random seeds and 1000 evaluation episodes.}
    \label{fig:mnist-sim-results}
\end{figure}

\noindent\textbf{Analysis.}
Figures \ref{fig:mnist-sim-results} shows that with enough training episodes in $\mathcal{D}$, ASE can learn a model of the simulated user that enables it to outperform a na\"{i}ve version of ASE that assumes the simulated user's belief update uses the same state encoder as the assistant's.
This result demonstrates that ASE can scale to training an expressive, recurrent neural network model of the user's belief update $b_{\theta}$.

\subsubsection{Scaling to Longer Observation Delays}

Our fifth experiment seeks to answer \textbf{Q5}: does ASE still improve the user's performance when observations are severely delayed?
We test this hypothesis with simulated users in the Car Racing domain from Section \ref{exp-known}.
We simulate the user using an expert policy trained via the model-based reinforcement learning method described in \citet{ha2018recurrent}.
In addition to manipulating the assistance condition as in Section \ref{exp-known}, we also manipulate the observation delay $d_{\mathrm{max}} \in \{0, 1, 2, ..., 20\}$.
The delay $d_{\mathrm{max}}$ controls the length of the no-delay and delay phases.
For example, in the user study in Section \ref{exp-known}, the no-delay and delay phases each lasted 5 timesteps, which corresponds to $d_{\mathrm{max}} = 5$.
In addition to measuring the task return, we also measure the simulated user's internal log-likelihood of the true state at each timestep.

\begin{figure}[t]
    \centering
    \includegraphics[height=0.35\linewidth]{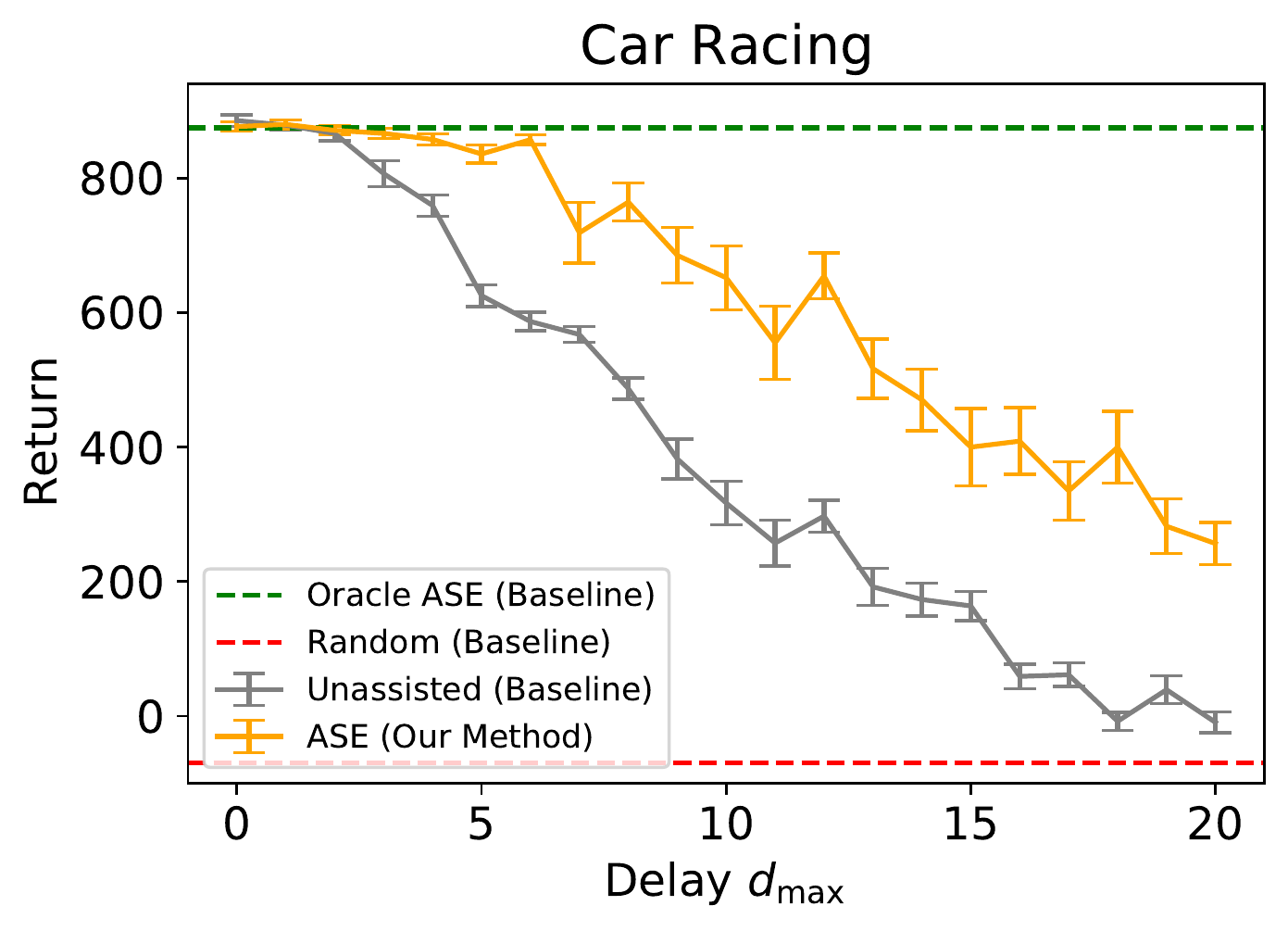}
    \includegraphics[height=0.35\linewidth]{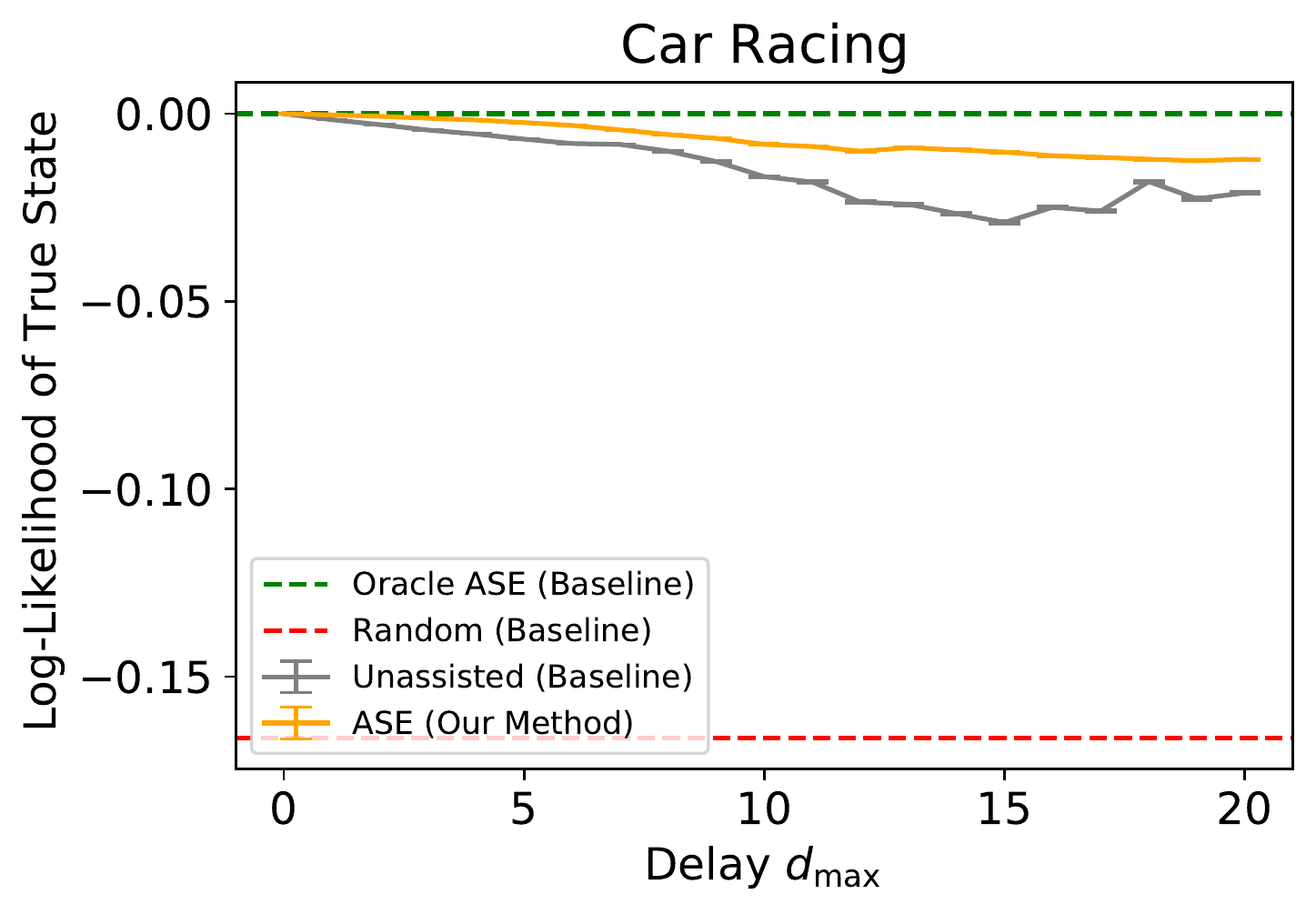}
    \caption{Car Racing experiments that address \textbf{Q5} -- does ASE still improve the user's performance when observations are severely delayed? -- by comparing our method (ASE), which tries to `undo' the observation delay $d_{\mathrm{max}}$ by predicting the current state and showing the user an observation representative of the predicted current state, to baselines that either show the human the outdated ambient observation generated by the environment (Unassisted) or randomly generate observations (Random). The results show that ASE substantially improves the simulated user's task performance (left plot) and the simulated user's internal state estimation accuracy (right plot), especially when the delay $d_{\mathrm{max}}$ is high. We measure standard error across 20 evaluation episodes. The gap between ASE and the oracle can be attributed to imperfections in the assistant's learned dynamics model, which is used to define its state encoder $f$.}
    \label{fig:car-sim-results}
\end{figure}

Figure \ref{fig:car-sim-results} shows that ASE substantially outperforms the unassisted and random baselines (orange vs. gray and red curves) in assisting simulated users.
ASE helps the simulated user by predicting the current state given outdated observations, then showing the user an observation representative of the predicted current state.
These predictions are not perfect, as shown by the gap between ASE and the oracle (orange vs. green curve), but still align the simulated user's beliefs more closely with the true state.
As the delay $d_{\mathrm{max}}$ increases, the assistant's dynamics model -- which is used to define its state encoder $f$ (see Section \ref{exp-known}) -- is not able to accurately predict the current state.
Hence, both assisted and unassisted performance decrease as the delay increases.

\subsection{Implementation Details} \label{imp-deets}

We use Adam \citep{kingma2014adam} to perform gradient descent on the objective in Equation \ref{eq:mle}.

\noindent\textbf{MNIST.}
There are $T = 28$ timesteps per episode, and the user can change their label at each timestep.
Each observation consists of zero or more row indices and corresponding pixel values: $\Omega = ([28] \times \mathbb{R}^{28})^\ast$, where $[28] = \{1, 2, ..., 28\}$ denotes the set of row indices, and $\ast$ denotes the Kleene star.
Let $\bI_{1:28, 1:28} \in \mathbb{R}^{28 \times 28}$ denote the full image.
The environment initially emits the full image observation $\bo_0 = ((1, \bI_{1,1}, \bI_{1,2}, ..., \bI_{1, 28}), (2, \bI_{2, 1}, ...), ..., (28, \bI_{28, 1}, ...))$, and the assistant observes it.
We define the assistant's belief update $b$ using Equation \ref{eq:continuous-belief-update}, where the state encoder $f$ is an LSTM sequence model  \citep{hochreiter1997long} trained to reconstruct the full image given a sequence of pixel observations.
We assume that the user's state estimation process lies in a singleton hypothesis space $\mathcal{B} = \{b_{\textbf{H}}\}$, where $b_{\textbf{H}}$ is identical to the assistant's belief update $b$, except that it ignores any observation that consists of more than one row of pixels.

Under these assumptions, the optimal synthetic observation $\tilde{\bo}_t$ consists of exactly one row of pixels.
Furthermore, $\tilde{\bo}_t$ minimizes the Euclidean distance between the user's latent state $f(\tilde{\bo}_{0:t-1}, \tilde{\bo}_t)$ after observing the partial image, and the assistant's latent state $f(\bo_0)$ after observing the full image (taking $\sigma^2 \to 0$ in the assistant's belief update in Equation \ref{eq:continuous-belief-update} simplifies the KL-divergence to the Euclidean distance between mean states).
To compute the assistant's belief state after observing the full image at time $t=0$, we break up the full image into an arbitrary sequence of pixel row observations, and feed each observation to the RNN state encoder $f$ one by one.
Note that all of this occurs immediately after the assistant observes the full image at time $t=0$: by the time the assistant computes the optimal observation for the user at time $t=0$, the assistant has already processed the entire sequence of observations.
In the simulation experiment in Appendix \ref{sim-exp-BI}, we train a policy $\pi_{\theta}$ (to model user actions via Equation \ref{eq:user-action}) end to end with the state estimation model $b_{\theta}$.

\noindent\textbf{Car Racing.}
There are a maximum of $T = 1000$ timesteps per episode.
Each observation consists of an image and a binary feature that indicates whether the image is delayed: $\Omega = \mathbb{R}^{64x64} \times \{0, 1\}$, where $0$ indicates no delay and $1$ indicates delay.
We define the assistant's belief update $b$ using Equation \ref{eq:continuous-belief-update}, where the state encoder $f$ is composed of a recurrent neural network (RNN) dynamics model \citep{schmidhuber1990making} and a variational auto-encoder model of image observations (VAE; \citealp{kingma2013auto}) trained on random trajectories \citep{ha2018recurrent}.
The state space $\mathcal{S}$ is the 256-dimensional latent space of the RNN $f$.
The VAE uses a 32-dimensional latent space to model 64x64 RGB image observations.
We assume that the user's state estimation process lies in a singleton hypothesis space $\mathcal{B} = \{b_{\textbf{H}}\}$, where $b_{\textbf{H}}$ is identical to the assistant's belief update $b$, except that it ignores the binary delay indicator in the observations and simply assumes all observations are not delayed.

Under these assumptions, the optimal synthetic observation $\tilde{\bo}_t$ minimizes the Euclidean distance between the user's latent state $f(\tilde{\bo}_{0:t-1}, \tilde{\bo}_t, \ba_{0:t-1})$ and the assistant's latent state $f(\bo_{0:t}, \ba_{0:t-1})$ (as in MNIST, taking $\sigma^2 \to 0$ in the assistant's belief update in Equation \ref{eq:continuous-belief-update} simplifies the KL-divergence to the Euclidean distance between mean states).
If the last $d$ observations are delayed, we approximate this solution using a prediction of a current, non-delayed observation: $\tilde{\bo}_t \leftarrow \hat{\bo}_t$.
This prediction is made by replacing the delayed observations $\bo_{t-d+1:t}$ with recursively predicted, non-delayed observations $\hat{\bo}_{i > t-d} = (g(f(\bo_{0:t-d}, \hat{\bo}_{t-d+1:i-1}, a_{0:i-1})), 0)$ from the RNN state encoder $f$ and VAE image decoder $g$, where the $0$ indicates that the predicted observation is not delayed.

\noindent\textbf{2D navigation.}
The states are arranged in a 5x5 grid, and the number of states is $|\mathcal{S}| = 100$.
Each state $(x, y, \phi)$ contains a discrete position $x, y \in \mathbb{N}$ and discrete orientation $\phi \in \{\mathrm{N}, \mathrm{S}, \mathrm{E}, \mathrm{W}\}$.
The actions, $\mathcal{A} = \{\text{turn left}, \text{turn right}, \text{move forward}\}$, change the user's orientation or position deterministically.
There are a maximum of $T = 25$ timesteps per episode.
The environment contains a discrete set of 78 objects (26 in each of the 3 categories): $\mathcal{X} = \{\text{chair, window, bathtub, painting, ...}\}$.
The observation space is the power set of the set of objects: $\Omega = \mathcal{P}(\mathcal{X})$.
The observation model $p^{\mathrm{obs}}(\bo|\bs)$ is a delta function on the subset of all objects $\bo$ visible from state $\bs$.
The assistant knows the locations of all objects, and can observe all objects in front of the user simultaneously: in other words, the assistant performs Bayesian belief updates (Equation \ref{eq:bayes-filter}) using the true observation model $p^{\mathrm{obs}}$.
We compute the optimal synthetic observation $\tilde{\bo}_t$ by simply enumerating the singleton sets of objects in $\Omega$ and computing the KL-divergence (Equation \ref{eq:induce}) for each possible value of $\tilde{\bo}_t$.

\noindent\textbf{Lunar Lander.}
We blend the color of the lander's body with the background to make it difficult for the user to see, making the tilt indicator more prominent.
Each episode ends when the lander contacts the ground, which typically occurs at $T = 115$ timesteps.
We define the assistant's belief update $b$ using Equation \ref{eq:continuous-belief-update}, where the state encoder $f$ simply passes through the most recent observation: $f(\bo_{0:t}, \ba_{0:t-1}) = \bo_t$.
We hardcode the optimal policy $\pi$ used to model user actions in Equation \ref{eq:user-action}: if the angle $s_t > 0$, then fire the right thruster to counter-rotate the vehicle counter-clockwise; or if the angle $s_t < 0$, then fire the left thruster to counter-rotate the vehicle clockwise.
To simplify the integral in Equation \ref{eq:user-action}, we take $\sigma^2 \to 0$ in the model of the user's belief update $b_{\theta}$ (Equation \ref{eq:continuous-belief-update}).

\noindent\textbf{Habitat navigation.}
The number of states is $|\mathcal{S}| = 1640$.
The number of objects is $|\mathcal{X}| = 34$.
The initial state distribution is uniform: $s_0 \sim \mathrm{Unif}(\mathcal{S})$.
At the beginning of the episode, the user has a uniform belief distribution over possible initial states.
For each episode, we sample a goal state uniformly at random from $\mathcal{S}$.
There are a maximum of $T = 100$ timesteps per episode.

\clearpage

\begin{table}
  \caption{Car Racing User Study}
  \label{tab:car-survey}
  \centering
  \small
  \begin{tabular}{llll}
    \toprule
    & $p$-value & Unassisted & ASE \\

I was able to keep the car on the road & $\mathbf{<.0001}$ & 1.67 & \textbf{3.75} \\
I could anticipate the consequences of my steering actions & $\mathbf{<.001}$ & 2.25 & \textbf{4.17} \\
I could tell when the car was about to go off road & $\mathbf{<.01}$ & 3.08 & \textbf{4.33} \\
I could tell when I needed to steer to keep the car on the road & $\mathbf{<.05}$ & 3.17 & \textbf{4.83} \\
I was often able to determine the car's current position & $\mathbf{<.05}$ & 3.50 & \textbf{4.75} \\
using the picture on the screen & & & \\
I could tell that the picture on the screen was sometimes delayed & $\mathbf{<.001}$ & 6.83 & \textbf{4.25} \\
The delay made it harder to perform the task & $\mathbf{<.01}$ & 6.58 & \textbf{4.83} \\

    \bottomrule
  \end{tabular}
  \captionsize
  \caption*{Subjective evaluations from 12 participants. Means reported below for responses on a 7-point Likert scale, where 1 = Strongly Disagree, 4 = Neither Disagree nor Agree, and 7 = Strongly Agree. $p$-values from a one-way repeated measures ANOVA with the presence of assistance as a factor influencing responses.}
\end{table}

\begin{table}
  \caption{2D Navigation User Study}
  \label{tab:gw-survey}
  \centering
  \small
  \begin{tabular}{lllll}
    \toprule
    & & $p$-value & Unassisted & Assisted \\

\multirow{4}{*}{\rotatebox[origin=c]{90}{Naive ASE}} & I was often able to infer my current position and orientation & $>.05$ & 5.50 & 5.67 \\
 & I was often able to move toward the goal & $>.05$ & 5.50 & 5.58 \\
 & I often found the guidance helpful & $>.05$ & 6.00 & 5.50 \\
 & I often forgot which position and orientation I believed was in & $>.05$ & 3.17 & 3.25 \\
\midrule
\multirow{4}{*}{\rotatebox[origin=c]{90}{ASE}} & I was often able to infer my current position and orientation & $\mathbf{<.01}$ & 5.50 & \textbf{6.83} \\
 & I was often able to move toward the goal & $\mathbf{<.01}$ & 5.50 & \textbf{6.83} \\
 & I often found the guidance helpful & $\mathbf{<.01}$ & 6.00 & \textbf{6.92} \\
 & I often forgot which position and orientation I believed was in & $\mathbf{<.01}$ & 3.17 & \textbf{1.42} \\

    \bottomrule
  \end{tabular}
  \captionsize
  \caption*{Subjective evaluations from 12 participants. Means reported below for responses on a 7-point Likert scale, where 1 = Strongly Disagree, 4 = Neither Disagree nor Agree, and 7 = Strongly Agree. $p$-values from a one-way repeated measures ANOVA with the presence of assistance as a factor influencing responses.}
\end{table}

\begin{table}
  \caption{Lunar Lander User Study}
  \label{tab:lander-survey}
  \centering
  \small
  \begin{tabular}{llll}
    \toprule
    & $p$-value & Unassisted & ASE \\

I could tell when the lander was tilted & $\mathbf{<.05}$ & 5.67 & \textbf{6.33} \\
I was able to straighten the lander before it tilted out of control & $>.05$ & 4.42 & 4.58 \\

    \bottomrule
  \end{tabular}
  \captionsize
  \caption*{Subjective evaluations from 12 participants. Means reported below for responses on a 7-point Likert scale, where 1 = Strongly Disagree, 4 = Neither Disagree nor Agree, and 7 = Strongly Agree. $p$-values from a one-way repeated measures ANOVA with the presence of assistance as a factor influencing responses.}
\end{table}

\begin{figure}[t]
    \centering
    \includegraphics[width=0.49\linewidth]{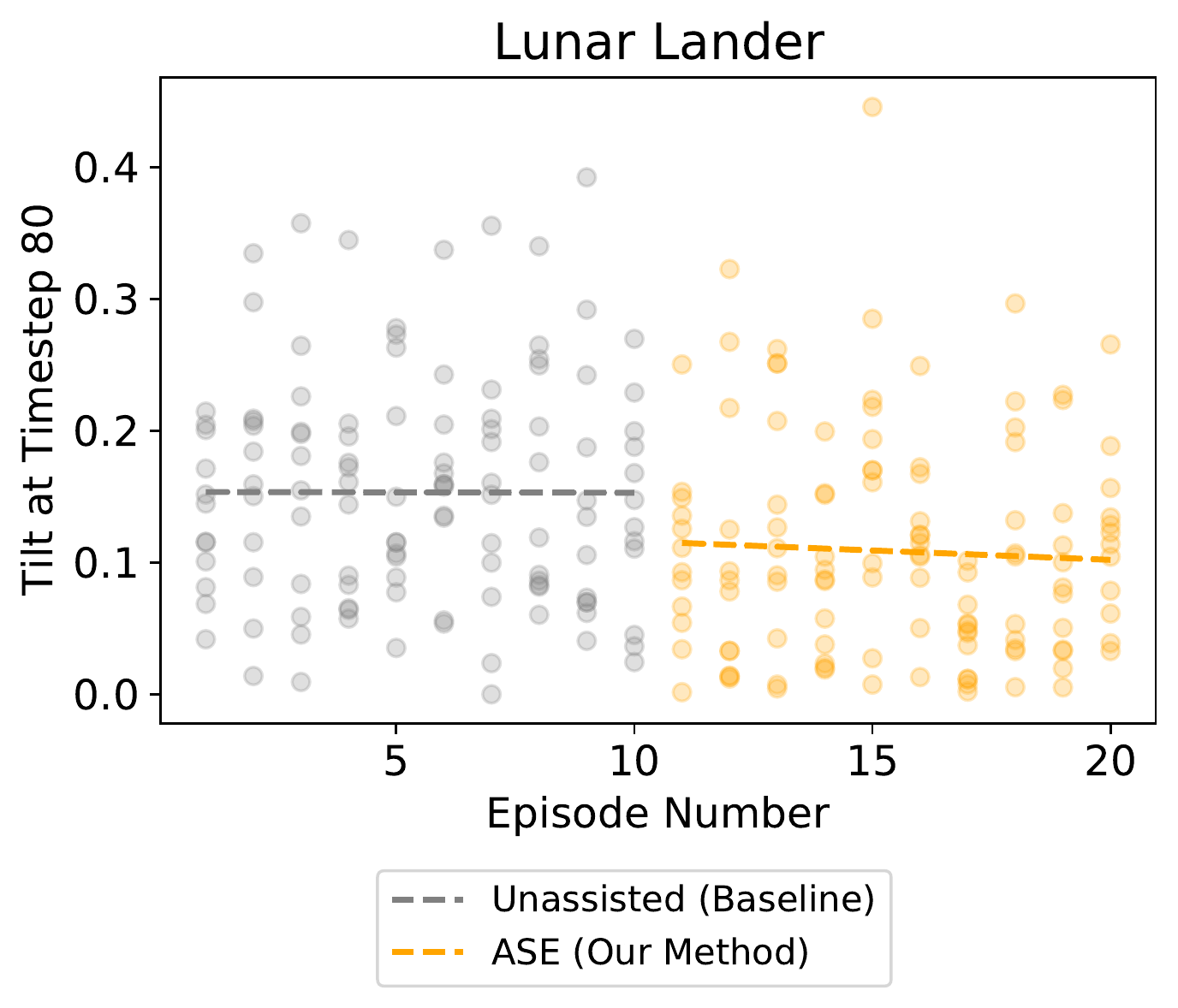}
    \includegraphics[width=0.49\linewidth]{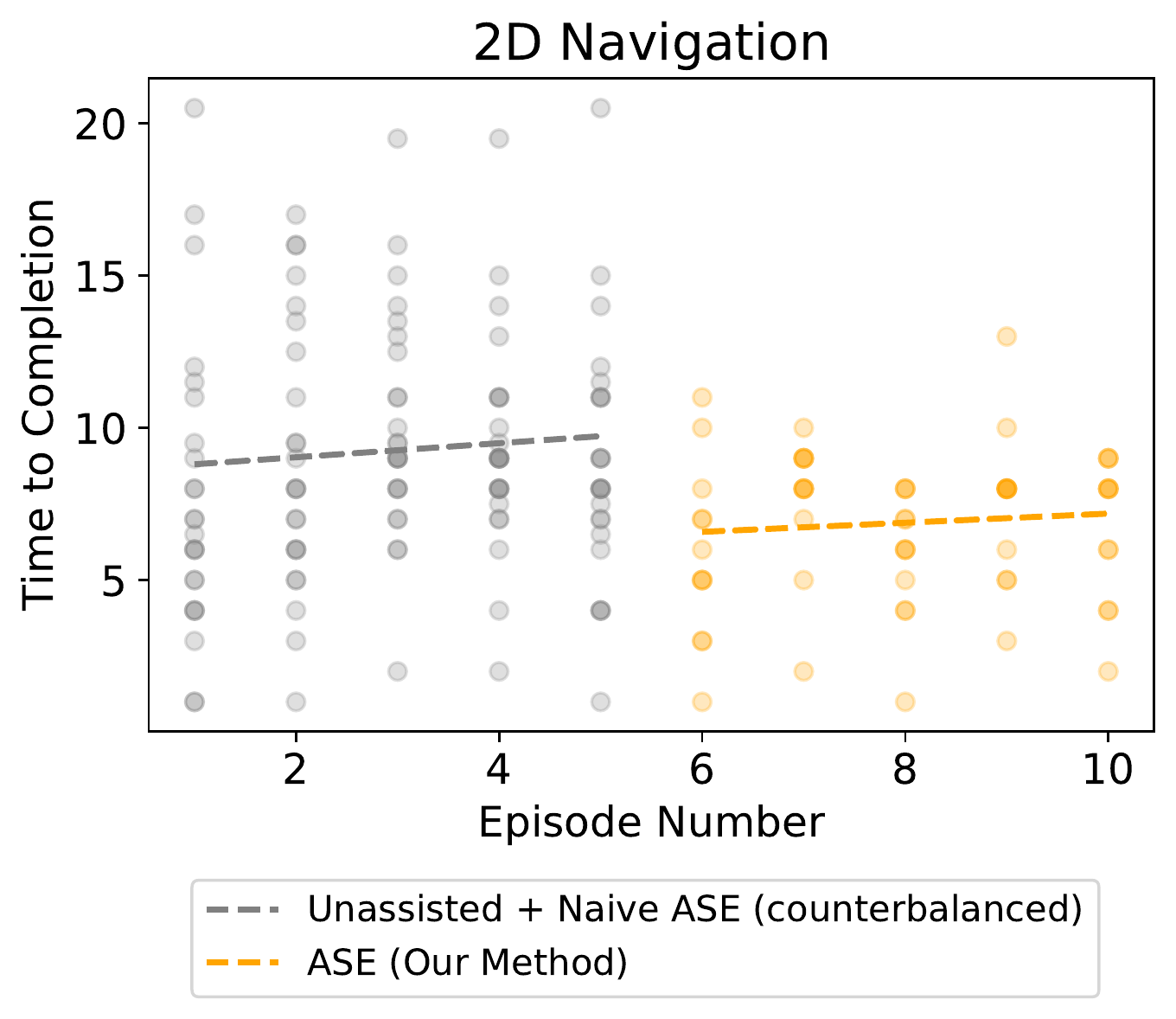}
    \caption{For a given episode number, each circle represents a different user. Each dashed line shows an ordinary least squares regression model trained on the data from a particular phase. Though we did not counterbalance the unassisted and ASE phases (only the unassisted and na\"{i}ve ASE phases), the learning effect does not appear to be a substantial confounder. Performance is relatively constant during the unassisted phase, and sharply improves once the ASE phase begins. This suggests that the improvement in performance between the unassisted and ASE phases is primarily due to the introduction of the ASE assistant, rather than a learning effect. We plot the tilt at timestep 80 for Lunar Lander, since that is when the performance improvements from assistance tend to appear (see plot (a) in Figure \ref{fig:lander-results}).}
    \label{fig:learning-effect}
\end{figure}

\end{document}